\documentclass[letterpaper, 10 pt, conference]{ieeeconf}

\IEEEoverridecommandlockouts                              % This command is only needed if 
\usepackage{graphics} % for pdf, bitmapped graphics files
\usepackage{epsfig} % for postscript graphics files
\usepackage{mathptmx} % assumes new font selection scheme installed
\usepackage{times} % assumes new font selection scheme installed
\usepackage{amsmath} % assumes amsmath package installed
\usepackage{amssymb}  % assumes amsmath package installed

\usepackage[english]{babel}
\usepackage{float}
\usepackage[letterpaper,top=2cm,bottom=2cm,left=3cm,right=3cm,marginparwidth=1.75cm]{geometry}
\usepackage[colorlinks=true, allcolors=blue]{hyperref}

\usepackage{subcaption}

\usepackage{lipsum}
\usepackage{comment}
\usepackage{xcolor}

\usepackage{url}

\definecolor{lightblue}{rgb}{0.5,.5,.8}
\graphicspath{{Figures/}}

\title{\LARGE \bf
Integrated Design and Control of a Robotic Arm on a Quadcopter for Enhanced Package Delivery*
}

\author{Animesh Singh$^{1}$, Jason Hillyer$^{1}$, Fariba Ariaei$^{2}$, and Hossein Jula$^{3}$ % <-this % stops a space
\thanks{*This work was not supported by any organization}% <-this % stops a space
\thanks{$^{1}$A.\ Singh and J.\ Hillyer are with the Department of Mechanical and Aerospace Engineering,
        University of California, Irvine, CA, USA. 
        {\tt\small animess3@uci.edu},{\tt\small hillyerj@uci.edu}}%
\thanks{$^{2}$F.\ Ariaei is with the Department of Mechanical and Aerospace Engineering,
        University of California, Irvine, CA, USA. 
        {\tt\small fariaei@uci.edu}}%
\thanks{$^{3}$H.\ Jula is with the Department of Electrical Engineering,
        California State University, Long Beach, CA, USA. 
        {\tt\small hossein.jula@csulb.edu}}%        
}

\begin{document}

\maketitle
\thispagestyle{empty}
\pagestyle{empty}

\begin{abstract}
This paper presents a comprehensive design process for the integration of a robotic arm into a quadcopter, emphasizing the physical modeling, system integration, and controller development. Utilizing SolidWorks for mechanical design and MATLAB Simscape for simulation and control, this study addresses the challenges encountered in integrating the robotic arm with the drone, encompassing both mechanical and control aspects. Two types of controllers are developed and analyzed: a Proportional-Integral-Derivative (PID) controller and a Model Reference Adaptive Controller (MRAC). The design and tuning of these controllers are key components of this research, with the focus on their application in package delivery tasks. Extensive simulations demonstrate the performance of each controller, with PID controllers exhibiting superior trajectory tracking and lower Root Mean Square (RMS) errors under various payload conditions. The results underscore the efficacy of PID control for stable flight and precise maneuvering, while highlighting adaptability of MRAC to changing dynamics.
\end{abstract}

\section{Introduction} \label{chp:intro}
Drones, or Unmanned Aerial Vehicles (UAVs), have recently gained significant attention for their numerous applications, such as aerial photography, package delivery, surveillance, mapping, environmental monitoring, and disaster response. Their maneuverability and accessibility to remote or hazardous environments make these systems particularly valuable. On the other hand, robotic arms have long been fundamental in various industries; however, their integration with drones remains under-explored due to inherent complexities and stability issues.

This research aims to explore the integration of robotic arms with drones to enhance their functional capabilities and advance automation in complex environments. In this paper, we present our research on modeling and controlling a quadcopter with an attached robotic arm using SolidWorks and MATLAB Simscape. This approach aims to reduce the cost and complexity of designing and implementing controllers for automation. The mechanical structure of the robotic arm was initially designed and developed in SolidWorks, then integrated into MATLAB Simscape and subsequently into the drone's model. The seamless integration of two separate assemblies is challenging and introduces model uncertainties. The combined Simscape model is used for designing model-based controllers and verification. Two control strategies—PID control and MRAC—were designed for automation and trajectory tracking. The performance of the system for trajectory tracking was compared.

The main contribution of this work lies in the physical modeling of complex systems to accurately represent their dynamics and interactions, their integration, and the design of robust and adaptive controllers for automation under conditions of uncertainty introduced by the integration of different systems and environments.

The remainder of this paper is structured as follows: Section \ref{chp:literatureReview} provides a literature review on modelling, designing, and controlling of drones and robotic arms. Section \ref{chp:method} outlines the methodologies used in our research. Section \ref{chp:simulator} discusses the design and control strategies employed. Section \ref{chp:results} presents the results and analysis, while Section \ref{chp:conclusion} concludes the paper and suggests future research.

\section{Literature Review} \label{chp:literatureReview}
Using MATLAB Simscape to simulate and analyze 3D models created from CAD software like SolidWorks is a well-established engineering practice. Various researchers have used these tools to enhance the understanding and performance of complex mechanical and robotic systems.

Cekus et al. \cite{article} have demonstrated comprehensive methodologies for creating simulation models through the integration of SolidWorks and MATLAB/Simulink. Their work focuses on modeling systems such as laboratory truck and forest cranes and designing control systems to perform motion analysis. Pozzi et al. \cite{article2} extend the capabilities of the SynGrasp Toolbox by integrating it with Simscape Multibody. Their work introduces new features such as gravity simulation, simultaneous grasping of multiple objects, environmental constraint inclusion, and advanced contact modeling and detection. Jatsun et al. \cite{Jatsun2020SynthesisOS} discuss using these tools to model a UAV quadcopter, implementing PID control strategies for dynamic simulations and trajectory planning. Mahto et al. \cite{mahto2022performance} and Long et al. \cite{Long2020} model and simulate robotic arms for industrial applications, utilizing SolidWorks for design and Simulink for control and motion analysis. Garcia et al. \cite{Garcia2021} and Pena et al. \cite{pena2020} explore the design and simulation of flexible systems and compliant mechanisms, employing Simscape to model dynamic responses, ensuring safety in human-robot interactions. Lee et al. \cite{Lee2018} explored the dynamic simulation of articulated robotic arms with embedded actuators. Other studies have similarly leveraged SolidWorks and MATLAB/Simulink for advanced robotic applications \cite{icscan2018design}, \cite{Urrea16,Urrea2020}, \cite{Eldirdiry2020}. 

Efforts to create flexible and open-source simulators have also gained traction. Guedelha et al. \cite{Guedelha2022} present a MATLAB/Simulink simulator for rigid-body articulated systems that supports various robotic configurations. Furthermore, Yura et al. \cite{Yura2017} cover the comprehensive modeling of a violin-playing robot arm using MATLAB/SimMechanics and integrating 3D CAD models from SolidWorks. Zhang et al. \cite{Zhang2019} present the kinematics simulation of a SCARA robot using MATLAB/SimMechanics.

Integrating SolidWorks with MATLAB Simscape proves effective for dynamic simulation and control analysis of mechanical and robotic systems. Contributions from various researchers demonstrate the utility of these tools in visualizing complex dynamics, improving control strategies, and enabling real-time simulations for diverse applications in engineering and robotics. This paper builds on these established methodologies to further explore the dynamic response of robotic systems, using Solidworks and MATLAB Simscape.

\section{Methodology and Tools} \label{chp:method}
In this section, we describe the mathematical and physical modeling of the robotic arm and its integration with the drone. The mathematical modeling of the arm is done using Lagrangian dynamics. The physical modeling of the entire system is done using SolidWorks and MATLAB Simscape to design an integrated system between the drone and the robotic arm. The Denavit-Hartenberg (DH) parameters and equations are used to calculate torques for the robotic arm joints. The entire simulation and control system design is carried out using MATLAB Simulink/Simscape.

\begin{figure}[thpb]
    \centering
    \parbox{3in}{\includegraphics[scale=0.28]{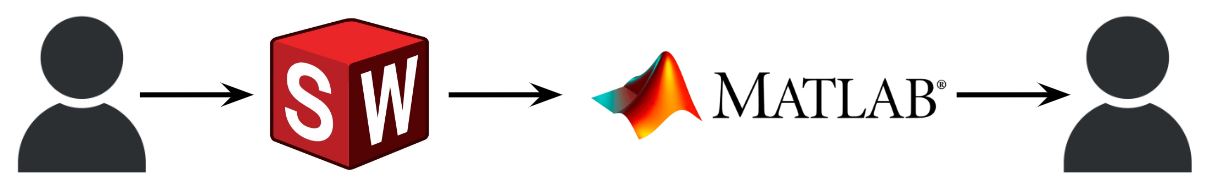}}
    \caption{Design Workflow}
    \vspace{-5mm}
%    \label{fig:enter-label}
\end{figure}

\subsection{Dynamical modeling of the robotic arm}
The theoretical analysis of the robotic arm is presented in the following subsection. We use Lagrangian dynamics to model the three-linkage robotic arm to drive the equations of motion.

\subsection{Lagrangian Dynamics}
We assume each arm link to be a thin cylindrical rod. Consequently, the inertia matrix for each link and the rotation matrix from each link body frame $\mathcal{B}_i$ to the inertial frame $\mathcal{O}$ are defined as follows:

\begin{equation*}
    I_i = m_i\begin{bmatrix}
            0 & 0 & 0\\
            0 & \frac{l_i^2}{12} & 0\\
            0 & 0 & \frac{l_i^2}{12}
        \end{bmatrix}
\end{equation*}
\begin{equation*}
    R_i^{\mathcal{O}} = \begin{bmatrix}
            \cos\left(\sum_{j=1}^{i} \theta_j\right) & -\sin\left(\sum_{j=1}^{i} \theta_j\right) & 0\\
            \sin\left(\sum_{j=1}^{i} \theta_j\right) & \cos\left(\sum_{j=1}^{i} \theta_j\right) & 0\\
            0 & 0 & 1
        \end{bmatrix}
\end{equation*}

where $m_i$ and $l_i$ are the mass and length of the $i$th link, respectively. We design a three-link robotic arm where $i \in \{1,2,3\}$. For simplicity, only planar motion is considered. Therefore, the angular velocity of each link is given by:
\begin{equation*}
    \vec{\omega}_i = \dot{q}_i\hat{k} = J_{\omega_i}\dot{q_i}
\end{equation*}

Here $q_i \in \mathbb{R}^3$, with $q_i(0)=0$, is the coordinate vector, and the $J_{\omega_i} \in \mathbb{R}^{3\times3}$ is a matrix with elements that are zero except for the last row of the first $i$ columns, which are ones. We assume that the system is initially at rest.

The velocities of the centers of mass for each link in the inertial frame can be expressed as:

\begin{equation*}
    \vec{v}_{c_i} = \sum_{j=1}^{i} \sum_{k=1}^{j} (2-\delta_{ij}) \frac{l_j}{2} \dot{q}_k
        \begin{bmatrix}
            -s_j \\
             c_j \\
             0
        \end{bmatrix}  
\end{equation*}
where $\delta_{ij}$ is the Kronecker delta function, and $s_j$ and $c_j$ represent sine and cosine functions defined as $s_j = \sin\left(\sum_{k=1}^{j} \theta_k\right)$ and $c_j = \cos\left(\sum_{k=1}^{j} \theta_k\right)$, respectively. Similar to the angular velocity, with an appropriate choice of a transfer matrix $J_{v_{c_i}} \in \mathbb{R}^{3\times3}$, the translational velocity can be written as $\vec{v}_{c_i} = J_{v_{c_i}} \dot{q_i}$. 

The kinetic energy $T$ can be formulated as:
\begin{equation*}
    T = \frac{1}{2}(m v^2 + I\omega^2)
\end{equation*}
where
\begin{equation*}
    \omega_i^2 = \vec{\omega}_i^T \vec{\omega}_i = \dot{q_i}^T J_{\omega_i}^T J_{\omega_i} \dot{q_i} 
\end{equation*}
\begin{equation*}
    v_{c_i}^2 = \vec{v}_{c_i}^T \vec{v}_{c_i} = \dot{q_i}^T J_{v_{c_i}}^T J_{v_{c_i}} \dot{q_i}
\end{equation*}

Thus, the kinetic energy for the robotic arm can be written as:
\begin{equation*}
    T = \sum_{i=1}^{3} \frac{1}{2} \dot{q}^T \left( m_i J_{v_i}^T J_{v_i} + J_{\omega_i}^T (R_i^{\mathcal{O}})^T I_i R_i^{\mathcal{O}} J_{\omega_i} \right) \dot{q_i}
\end{equation*}

The potential energy $V$ can be expressed as:
\begin{equation*}
   V = m_1 g \frac{l_1}{2} + \sum_{i=2}^{3} \sum_{j=1}^{i} m_i g (2-\delta_{ij}) \frac{l_j}{2} s_j
\end{equation*}
and consequently, the Lagrangian can be defined as:
\begin{equation*}
    L = T - V
\end{equation*}
The equations of the kinetic and potential energies can be substituted in the Lagrange's equation as follows
\begin{equation}
    \frac{d}{dt}\left(\frac{\partial L}{\partial \dot{q}_i}\right) - \frac{\partial L}{\partial q_i} = Q_i
\end{equation}
where $Q_i$ represents the generalized forces, which are particularly significant at the end of the third link where an end effector is present. Solving this gives us the required equations of motion for the arm.
%These equations, incorporating both kinetic and potential energies, enable us to predict the motion of the arm by solving the second-order differential equations for the generalized coordinates, $q_i$. With the Lagrangian mechanics framework in place, we can now systematically explore the system's responses to different inputs, optimize its performance, and enhance its stability and accuracy in performing complex tasks.

\subsection{Denavit Hartenberg Table and Parameters}
To describe the position and orientation of each link of the robotic arm, we use the DH convention \cite{niku2020introduction}. This method standardizes the representation of coordinate frames using four parameters: link length, link twist, link offset, and joint angle. We assume the motion of the arm is entirely planar, and the orientation of the claw is independent of the kinematics of the three arm links. Thus, the kinematics and inverse kinematics of the end effector can be solved independently of the arm links themselves to determine the forward and inverse kinematics solutions.

The DH Table for the 3-link arm has been assembled below 
\begin{table}[!h]
\centering
\caption{DH Table for the Robotic Arm}
\begin{tabular}{|| c | c | c | c | c ||} 
 \hline
 Joint i & $\theta_i$ & $d_i$ & $\alpha_{i,i+1}$ &$ a_{i,i+1}$\\ [0.5ex] 
 \hline\hline
 1 & $\theta_1$ & 0 & $\pi/2$ & 0\\ 
 \hline
 2 & $\theta_2$ & 0 & 0 & $l_1$\\
 \hline
 3 & $\theta_3$ & 0 & 0 & $l_2$\\
 \hline
 4 & $\theta_4$ & 0 & $\pi/2$ & $l_3$\\
 \hline
 5 & $\phi_1$ & 0 & $\pi/2$ & 0\\
 \hline
 6 & $\phi_2$ & 0 & $-\pi/2$ & 0\\
 \hline
 7 & $\phi_3$ & 0 & 0 & 0\\
 \hline
\end{tabular}
\end{table}\\
\noindent
where,

\noindent
$\theta_i$: Angle between $x_{i-1}$ and $x_i$ axes about $z_{i-1}$\\
$d_i$: Distance between $x_{i-1}$ and $x_i$ axes along $z_{i-1}$\\
$\alpha_{i,i+1}$: Angle between $z_{i-1}$ and $z_i$ axes along $x_{i}$\\
$a_{i,i+1}$: Distance between $z_{i-1}$ and $z_i$ axes along $x_{i-1}$\\
$d_1$: Distance between base of drone and first link.\\
$l_1$: Length of first link.\\
$l_2$: Length of second link.\\
$l_3$: Length of third link.\\

\begin{figure}[t!]
\centering
\begin{subfigure}[t]{.1\linewidth}
    \centering
    \includegraphics[width=0.8\linewidth]{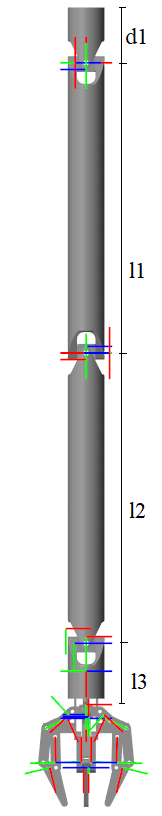}  
    \caption{Frames}
%    \label{fig:enter-label}
\end{subfigure}
\hfill
\begin{subfigure}[t]{.35\linewidth}
    \centering
    \includegraphics[height=0.6\textwidth]{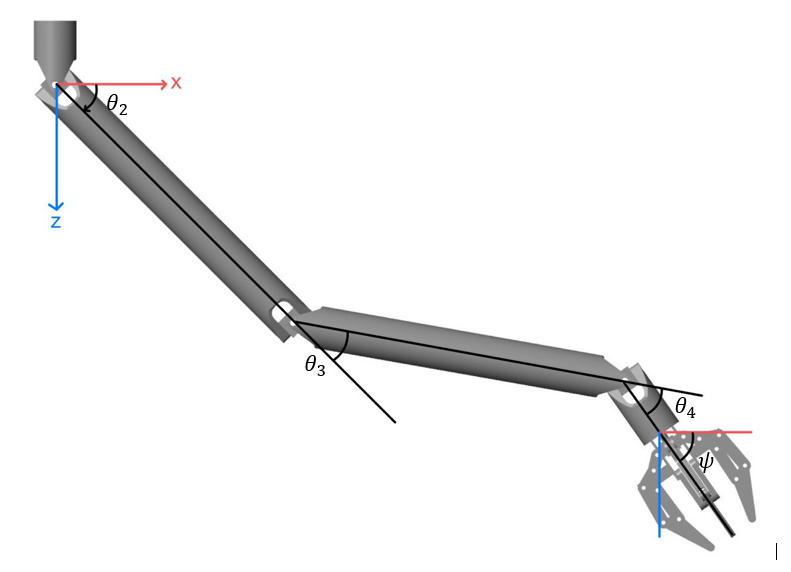}
    \caption{Angles}
%    \label{fig:enter-label}
\end{subfigure}
\begin{subfigure}[t]{.35\linewidth}
    \centering
    \includegraphics[height=0.4\textwidth]{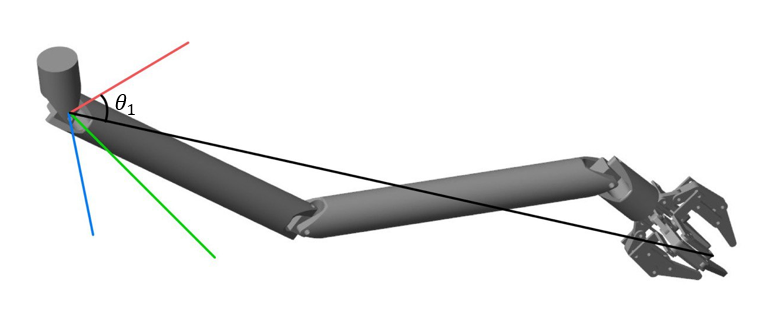}
    \caption{Top view}
%    \label{fig:enter-label}
\end{subfigure}
\caption{Arm's parameters for the DH frame}
\vspace{-5mm}
\label{fig: components}
\end{figure}

The displacement matrices used to evaluate the DH table are $D_x(\alpha,a)$ and $D_z(\gamma,c)$, representing displacement by $a$ and rotation by $\alpha$ along the $x$ direction and displacement by $c$ and rotation by $\gamma$ along the $z$ direction, respectively.
\begin{equation*}
    D_x(\alpha,a) = \begin{bmatrix}
        1 & 0 & 0 & a &\\
        0 & \cos\alpha & -\sin\alpha &0\\
        0 & \sin\alpha & \cos\alpha &0\\
        0 & 0 & 0 & 1
    \end{bmatrix}
\end{equation*}
\begin{equation*}
    D_z(\gamma,c) = \begin{bmatrix}
        \cos\gamma & -\sin\gamma & 0 & 0 \\
        \sin\gamma & \cos\gamma & 0 &0\\
        0 & 0 & 1 & c\\
        0 & 0 & 0 & 1
    \end{bmatrix}
\end{equation*}
\indent The kinematic equations are derived by successive multiplication of \( D_z \) and \( D_x \) matrices as per the DH table. Given the independence of the end effector kinematics from the arm links' kinematics, \( K \) can be expressed as:
\[
K = RE
\]
where
\begin{align*}
    R &= D_z(\theta_1, 0) D_x(\pi/2, 0) D_z(\theta_2, 0) D_x(0, l_1) \\
      &\quad D_z(\theta_3, 0) D_x(0, l_2) D_z(\theta_4, 0) D_x(\pi/2, l_3)
\end{align*}
\begin{align*}
    E &= D_z(\phi_1, 0) D_x(\pi/2, 0) D_z(\phi_2, 0) D_x(-\pi/2, 0) \\
      &\quad D_z(\phi_3, 0) D_x(0, 0)
\end{align*}
The $R \in \mathbb{R}^{4 \times 4}$ is called the reach matrix, representing the kinematics of the arm links, and $E$ is the end effector matrix. To solve the inverse kinematics problem, we define the task matrix \( T \) as:
\[
T = \begin{bmatrix}
a & b & c & x \\
d & e & f & y \\
g & h & i & z \\
0 & 0 & 0 & 1
\end{bmatrix}
\]
For the arm to reach the task position, the difference \( T - K \) must be zero, thus \( T = K \). The fourth column of \( T \) corresponds to the desired coordinates of the end of the third arm link.
The fourth column of \( K \) is equal to $(L \cos\theta_1, L \sin\theta_1, S, 1)^T$, where, $L = \sum_{i=1}^{3} l_i \cos(\sum_{j=1}^{i} \theta_{j+1})$ and $S = \sum_{i=1}^{3} l_i \sin(\sum_{j=1}^{i} \theta_{j+1})$.

To determine the joint angles, the value of \(\theta_1\) is derived by:
\begin{equation*}
\theta_1 = \arctan\left(\frac{y}{x}\right)
\end{equation*}
Next, for the azimuthal angle \(\psi\), the position of the 3rd joint is given by:
\begin{equation*}
p_{3x} = x - l_3 \cos \psi, \quad p_{3z} = z + l_3 \sin \psi
\end{equation*}
\(\theta_3\) is found using:
\begin{equation*}
\theta_3 = \pm \arccos\left( \frac{p_{3x}^2 + p_{3z}^2 - l_1^2 - l_2^2}{2 l_1 l_2} \right)
\end{equation*}
Let \(\eta = \arctan\left( \frac{p_{3z}}{p_{3x}} \right)\), and then solve for \(\theta_2\):
\begin{equation*}
\theta_2 = \eta - \arctan\left( \frac{l_1 + l_2 \cos \theta_3}{l_2 \sin \theta_3} \right)
\end{equation*}
Finally, \(\theta_4\) is determined from \(\psi\):
\begin{equation*}
\theta_4 = \psi - (\theta_2 + \theta_3)
\end{equation*}
To find the end effector kinematics, solve for \(E\) in terms of the task matrix \(T\) and the reach matrix \(R\):
\begin{equation*}
E = R^{-1} T
\end{equation*}
These steps complete the inverse kinematics analysis, facilitating the calculation of joint angles required to achieve a specified end-effector position.

\subsubsection{Control of the Arm}
The joint torques are input as described by the following function:
\begin{equation*}
\tau = \tau_{final}(1 - e^{-t/5})
\end{equation*}
where $\tau$ is the input torque to a joint and $\tau_{final}$ is the peak torque required to rotate the joint and arm link to the desired final position.
This function generates a time series, representing an increasing function of torque with time for each joint. This time series is then reversed, and the decreasing torques are subsequently applied to the joint to bring it back to its equilibrium position. Hence,
\begin{equation*}
\tau_{in} = \begin{bmatrix} \tau & \tau_{reverse} \end{bmatrix}
\end{equation*}

\begin{figure}[t!]
\centering
\includegraphics[width=0.52\textwidth]{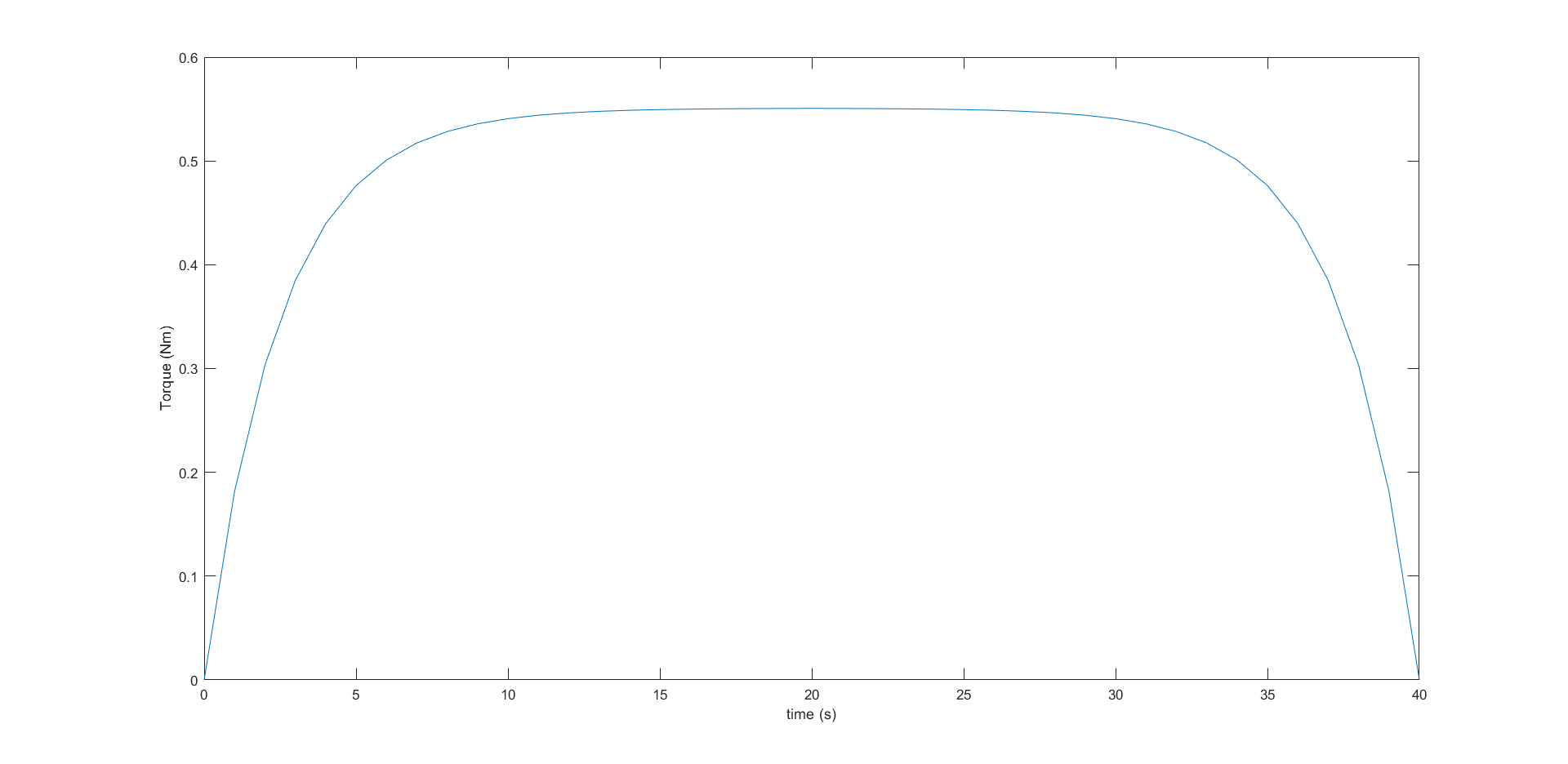}
\caption{Torque function example}
\vspace{-5mm}
%\label{fig:enter-label}
\end{figure}

\subsubsection{Example}
\begin{figure}[t!]
    \centering
    \includegraphics[width=0.48\textwidth]{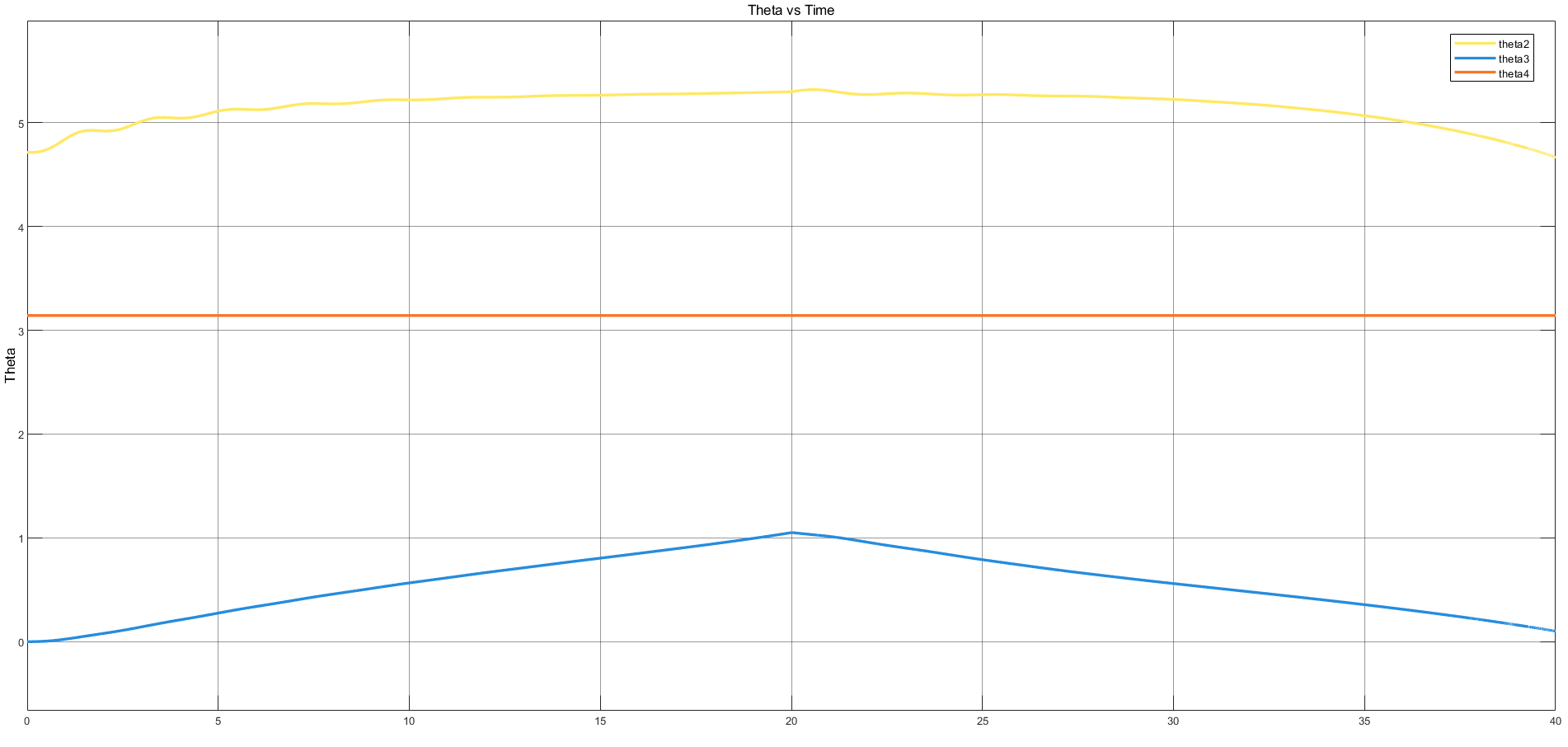}
    \caption{Variation of Joint Angles (rad) with time}
    \label{fig:exampleResultsAngles}
\end{figure}

\begin{figure}[t!]
    \centering
    \begin{subfigure}[t]{0.4\linewidth}
        \centering
        \includegraphics[width=12mm]{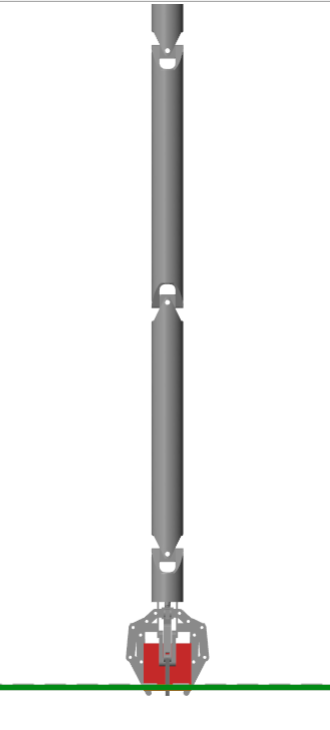}
        \caption{t = 0s}
        \label{fig:armcontrol1}
    \end{subfigure}
    \begin{subfigure}[t]{0.4\linewidth}
        \centering
        \includegraphics[width=15mm]{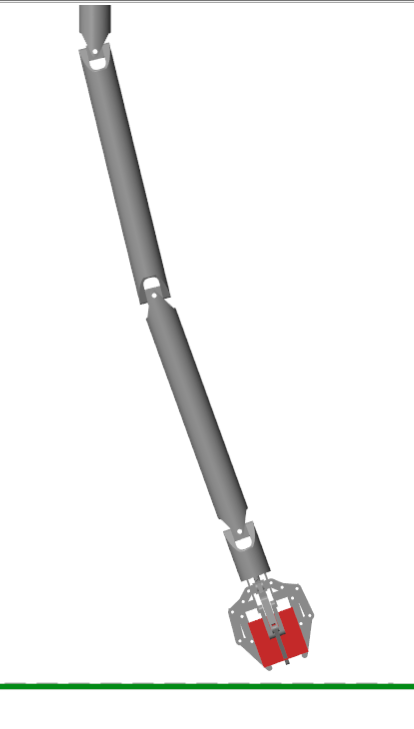}
        \caption{t = 2.5s}
        \label{fig:armcontrol2}
    \end{subfigure}
    \vfill
    \begin{subfigure}[t]{0.4\linewidth}
        \centering
        \includegraphics[width=24mm]{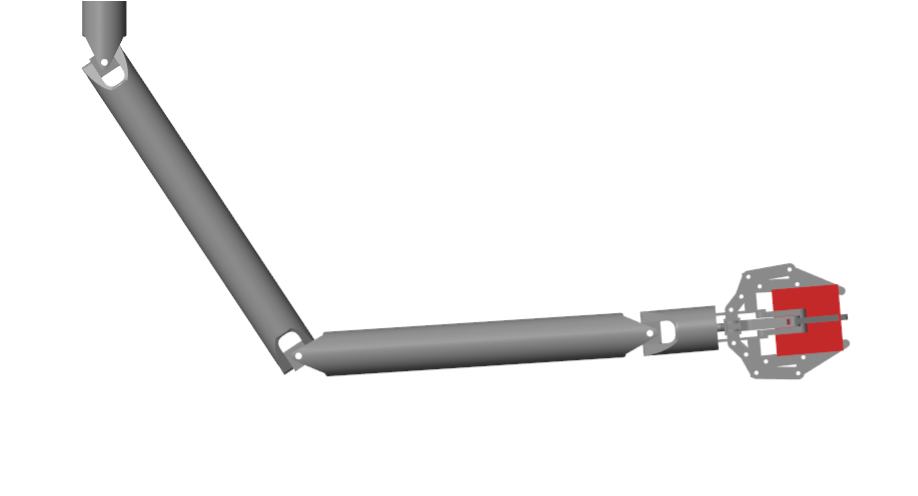}
        \caption{t = 20s, arm reaches desired position}
        \label{fig:armcontrol3}
    \end{subfigure}
    \begin{subfigure}[t]{0.4\linewidth}
        \centering
        \includegraphics[width=18mm]{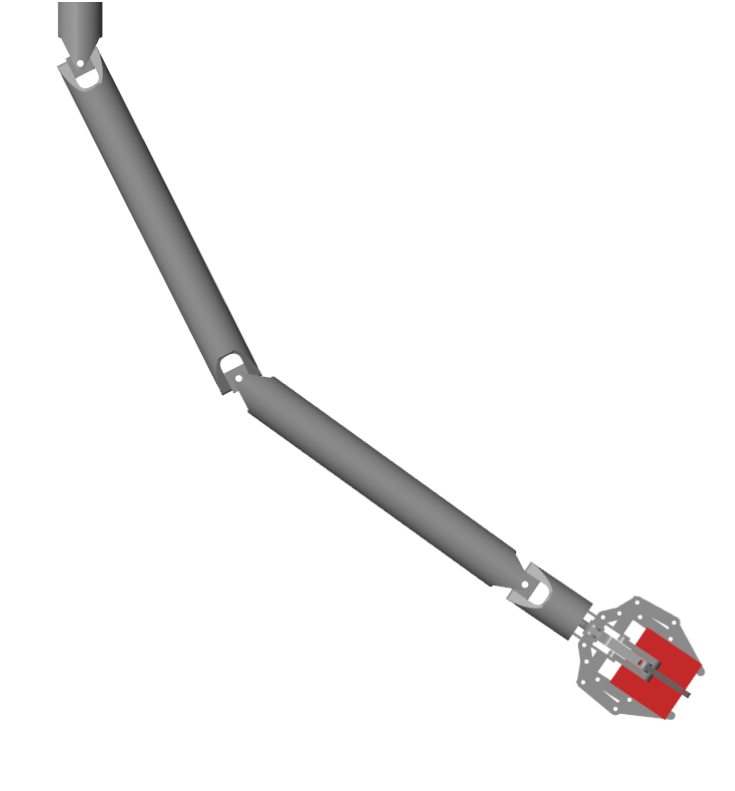}
        \caption{t = 30s, arm returning to rest position}
        \label{fig:armcontrol4}
    \end{subfigure}
    \caption{Timeline of the motion of the arm}
    \vspace{-5mm}
    \label{fig:armcontrol}
\end{figure}

The desired $(x, y, z)$ coordinates for the end effector are set to $(0.9, -0.38, 0)$ $(m)$, with an orientation of zero degrees. This configuration yields the following theoretical joint angles at the desired position:
\begin{equation*} \theta_1 = 0^\circ, \quad \theta_2 = 300^\circ, \quad \theta_3 = 60^\circ, \quad \theta_4 = 180^\circ \end{equation*}
\indent These values are found using DH table calculations shown in the previous section. The payload is a box with dimensions $8 \times 8 \times 8$ $(cm^3)$ and a density of 1000 $(kg/m^3)$, resulting in a weight of $W_\text{load} = 5.02$ $(N)$.
This load is held by the end effector, which also accounts for the weight of the arm links.
The simulation results are depicted in Fig. \ref{fig:exampleResultsAngles} and \ref{fig:armcontrol1}. The joint angles are measured as:
\begin{equation*} \theta_{m_1} = 0^\circ, \hspace{0.1cm} \theta_{m_2} = 303^\circ,  \hspace{0.1cm} \theta_{m_3} = 60.21^\circ,  \hspace{0.1cm} \theta_{m_4} = 180^\circ \end{equation*}
where $\theta_m$ denotes the measured angle.

\section{Physical Modeling and Simulator Design} \label{chp:simulator}
\subsection{Design with SolidWorks}
The first phase of the project involved designing the 3-link robotic arm using SolidWorks. The robotic arm was designed to accommodate the carrying capacity of one of the largest commercially available utility quadcopters, the DJI Matrice 300 RTK, which is approximately $2.7$ $(kg)$. The dimensions and materials were scaled and selected for aesthetics, durability, and weight constraints. Delrin plastic was specified as the material for the robotic arm in SolidWorks due to its strength and ease of fabrication. The final weight of the robotic arm was determined to be $1.92$ $(kg)$, allowing for a payload carrying capacity of $0.78$ $(kg)$ for the entire system.

\begin{figure}[t!]
\centering
\begin{subfigure}[t]{.3\linewidth}
\centering
\includegraphics[width=1\linewidth]{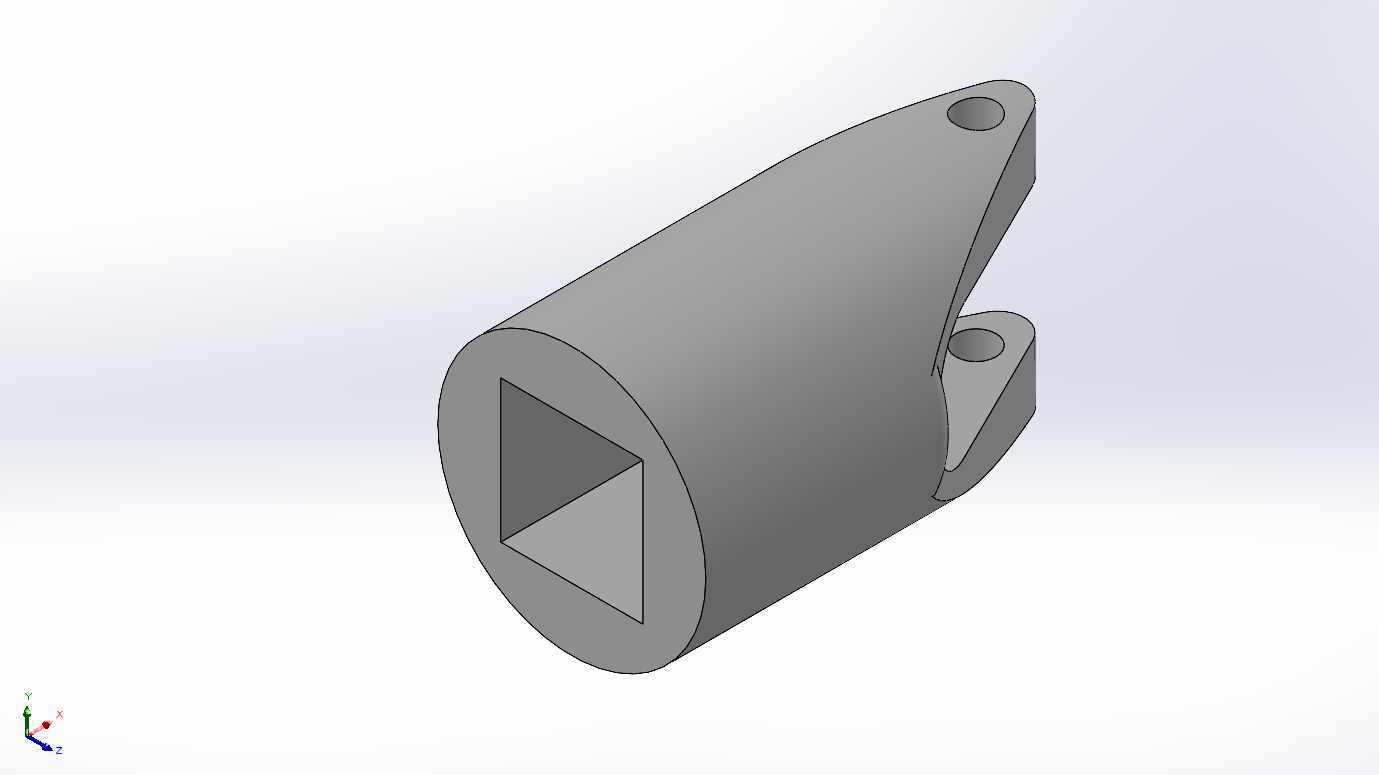}
\caption{Rotating base} \label{fig:RotatingBase}
\end{subfigure}
\begin{subfigure}[t]{.3\linewidth}
\centering
\includegraphics[width=1\linewidth]{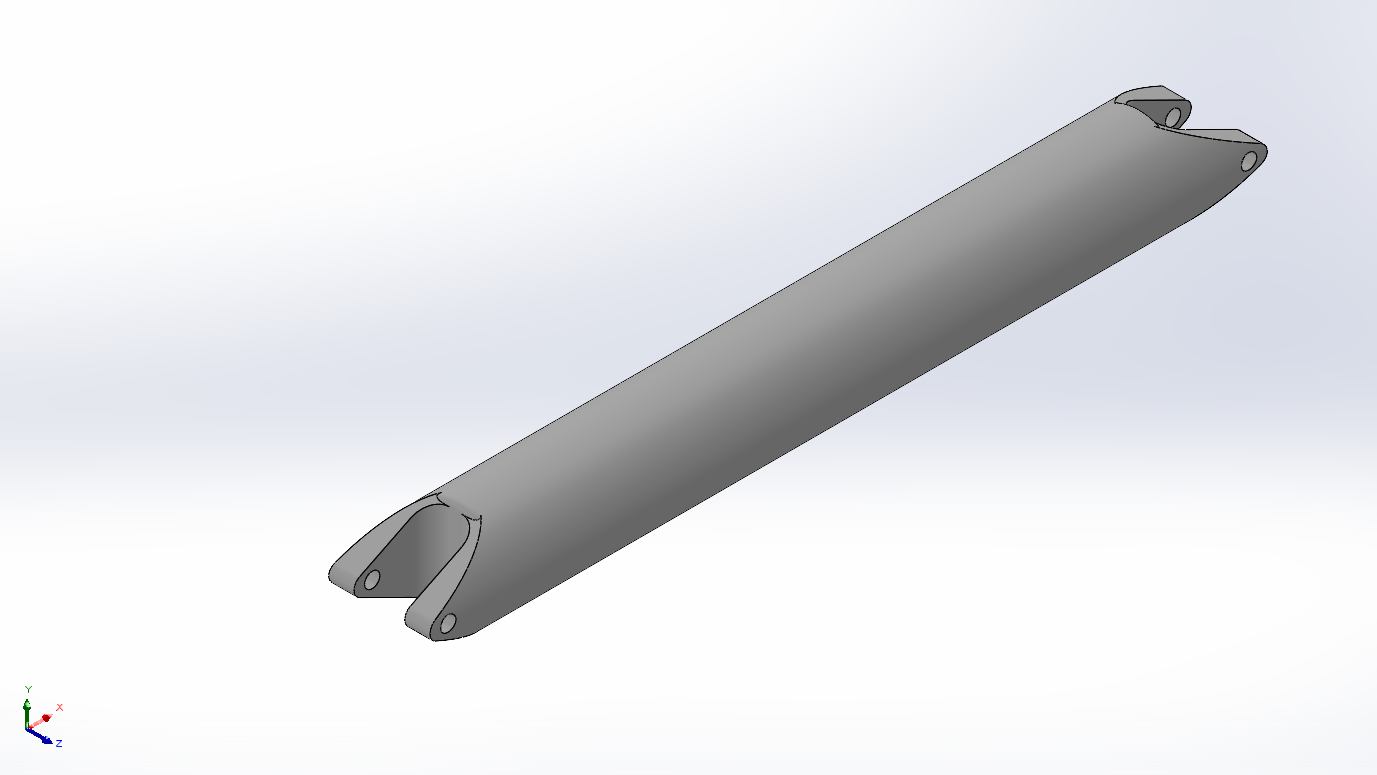}
\caption{Main linkage }
\label{fig:MainLinkage}
\end{subfigure}
\begin{subfigure}[t]{.3\linewidth}
\centering
\includegraphics[width=1\linewidth]{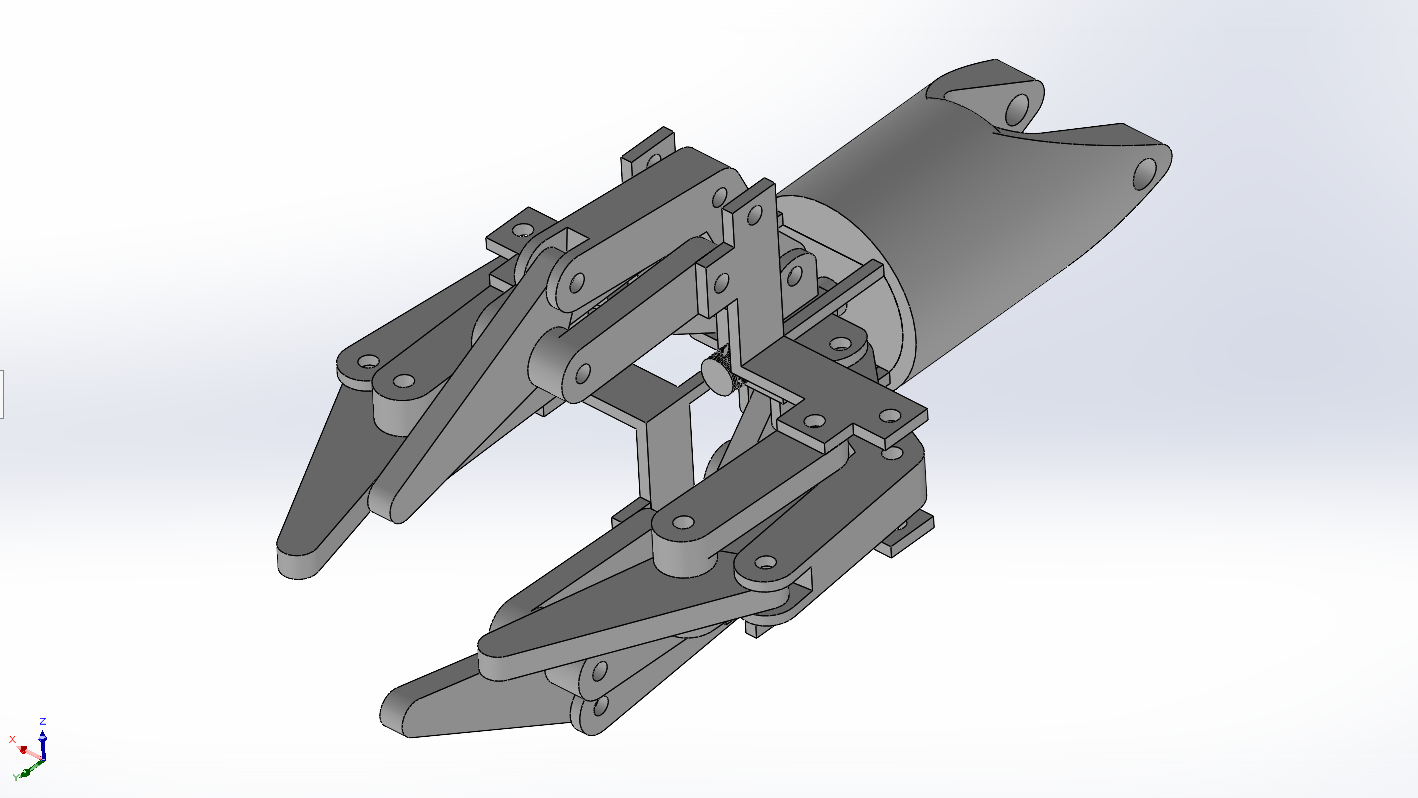}
\caption{Gripper} \label{fig:Gripper}
\end{subfigure}
\caption{Components of the robotic arm} \label{fig:components}
\end{figure}

Fig. \ref{fig:components} shows the components of the robotic arm. Fig. \ref{fig:RotatingBase} displays the rotating base that connects the twin linkages to the quadcopter, featuring the same connection geometry as the twin linkages. Fig. \ref{fig:MainLinkage} shows the main linkage, of which there are two in the robotic arm assembly. This design facilitates simple fabrication and assembly. One linkage attached to a rotating base allows for hemispherical actuation to a distance equal to the length of the linkage from the rotating base. Adding a second linkage extends the range of actuation and allows access to regions closer to the rotating base that a single linkage could not reach on its own. By attaching this system of a rotating base and two twin linkages to a quadcopter, the arm can reach any position below the quadcopter. Fig. \ref{fig:Gripper} illustrates the gripper component attached to the extremity of the twin linkage system, designed to be actuated with a servo motor.

\begin{figure}[t!]
\centering \includegraphics[width=0.48\textwidth]{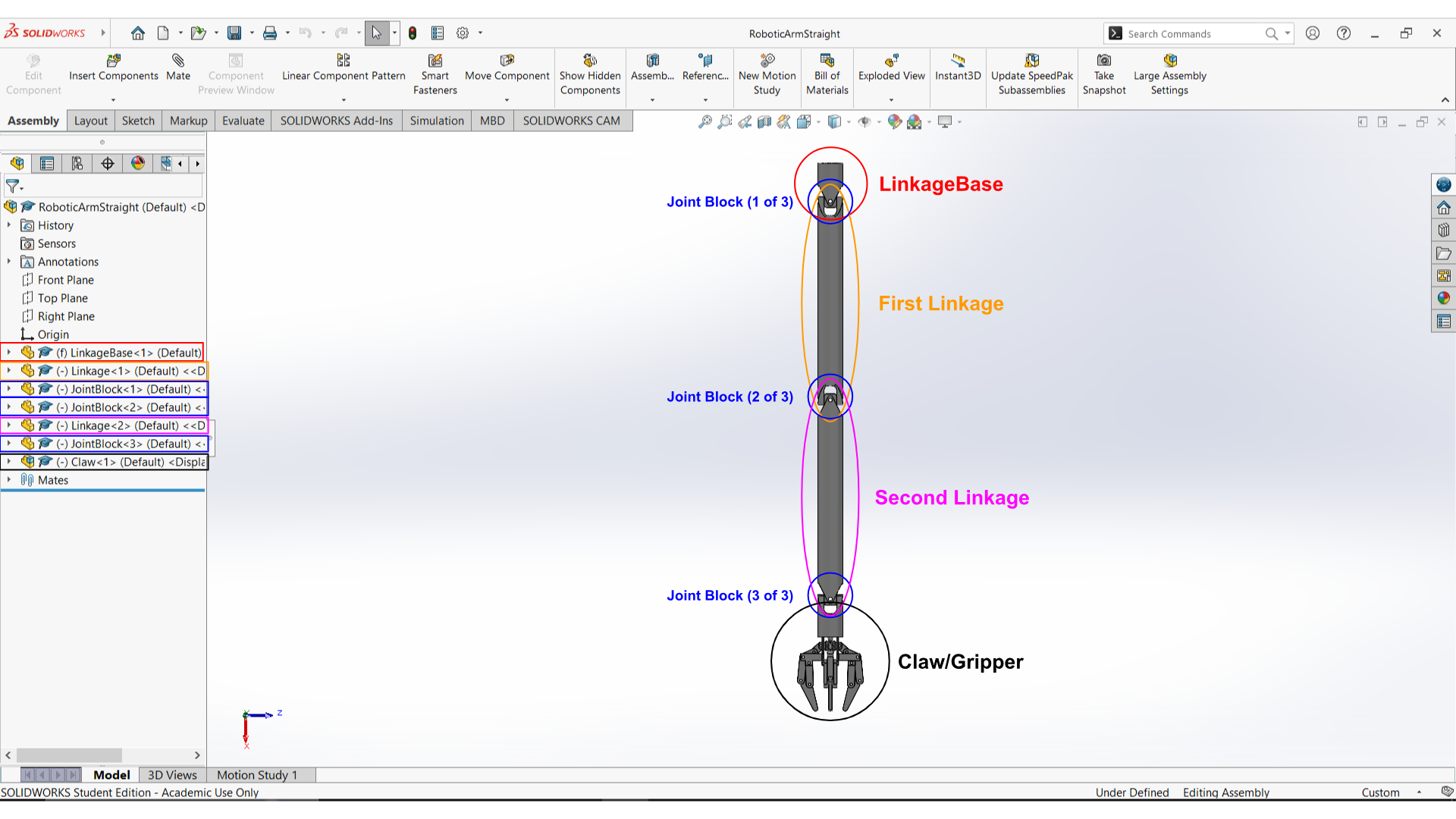}
\caption{The robotic arm designed using SolidWorks}
\vspace{-5mm}
\label{fig:solidworks-arm}
\end{figure}

The resulting SolidWorks assembly in Fig. \ref{fig:solidworks-arm} served as a virtual prototype of the robotic arm.

\subsection{Simulations with MATLAB Simscape }

To test dynamic scenarios, the SolidWorks model was imported into MATLAB Simscape. This integration allows direct connection between the CAD model and the simulation environment for analyzing the arm's behavior under various conditions.

\begin{figure}[t!]
    \centering
    \includegraphics[width=0.48\textwidth]{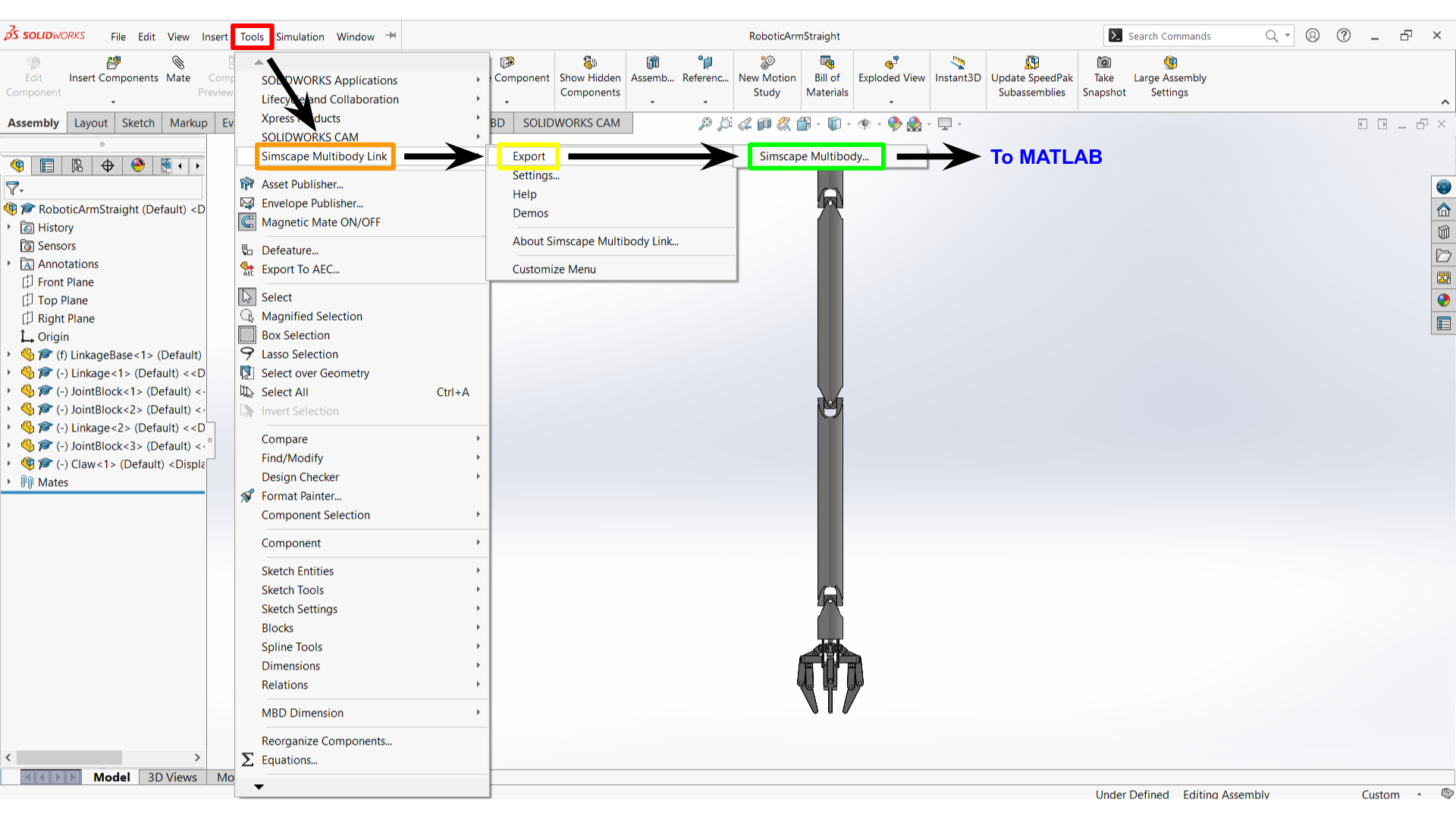}
    \caption{Exporting CAD model from SolidWorks to MATLAB Simscape}
    \vspace{-3mm}
    \label{fig:solidworks-matlab}
\end{figure}

\begin{figure}[t!]
    \centering  \includegraphics[width=0.48\textwidth]{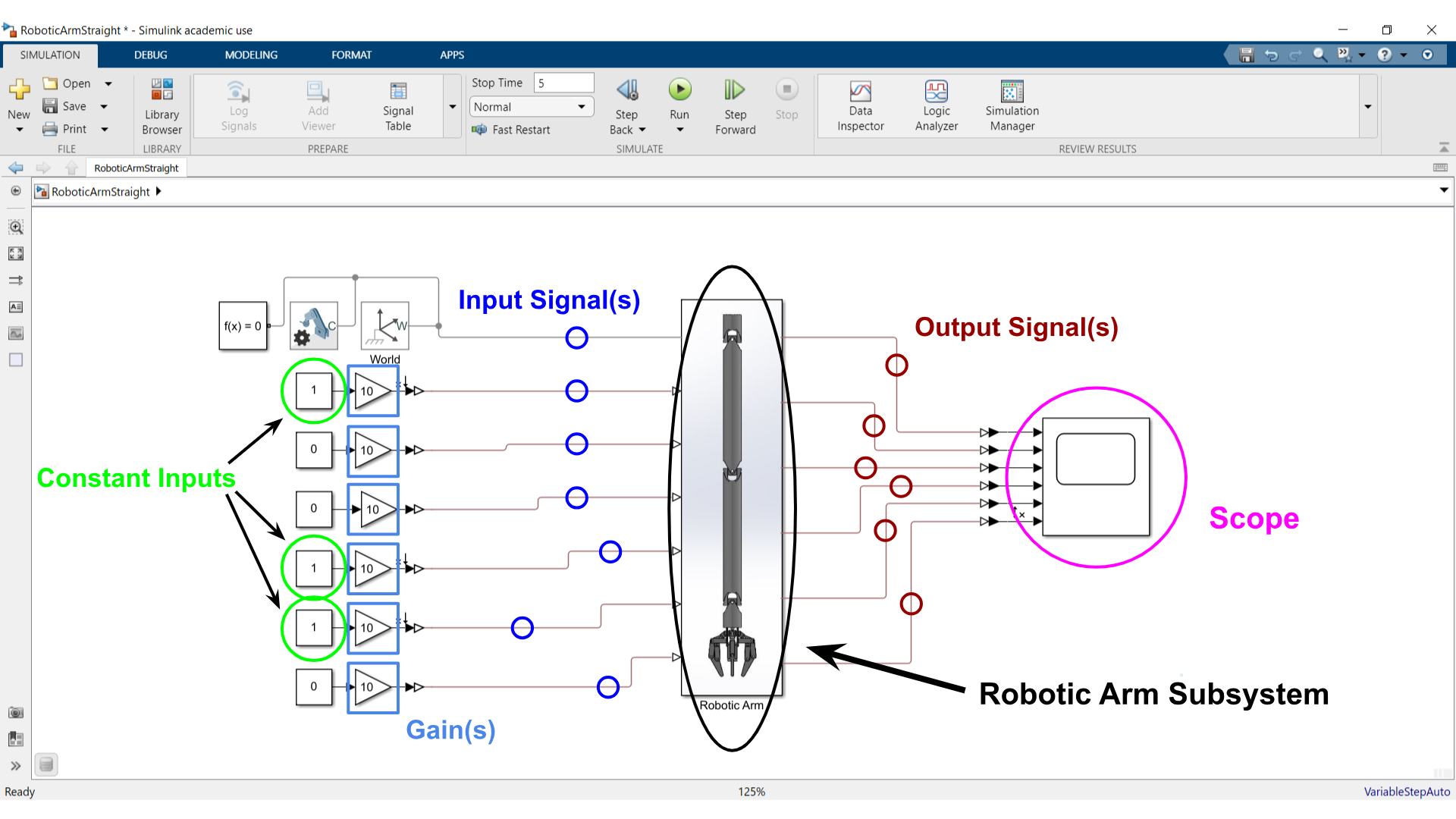}
    \caption{Simulink model organized with subsystems}
    \label{fig:SimscapModelSubsystems}
\end{figure}

Upon importing the model, Simscape automatically generates a corresponding model with predefined joints and constraints from SolidWorks.
In Simscape, blocks can be grouped and labeled to enhance user-friendliness. Fig. \ref{fig:SimscapModelSubsystems} shows the Robotic Arm Subsystem, with all relevant blocks grouped and labeled with a picture from SolidWorks.
\begin{figure}[t!]
    \centering \includegraphics[width=0.48\textwidth]{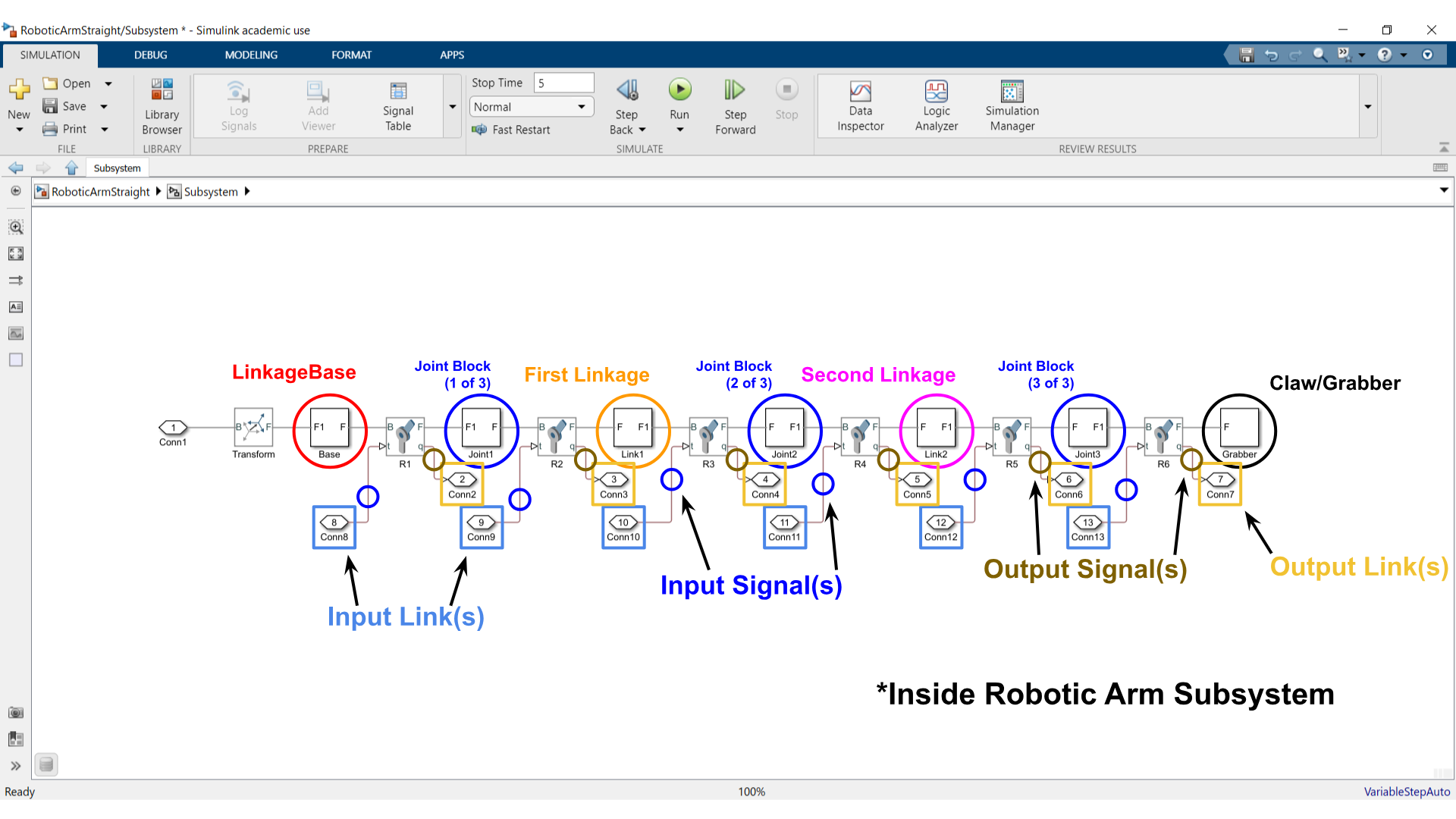}
    \caption{View inside the robotic arm subsystem in MATLAB Simscape}
    \vspace{-5mm}
    \label{fig:subsytem}
\end{figure}
Once the functionality is determined, MATLAB Simscape can run dynamic simulations. This allows analysis of the robotic arm's response to various inputs and external forces and refining the control scheme by observing the arm's response.

\begin{figure}[t!]
    \centering
    \includegraphics[width=0.48\textwidth]{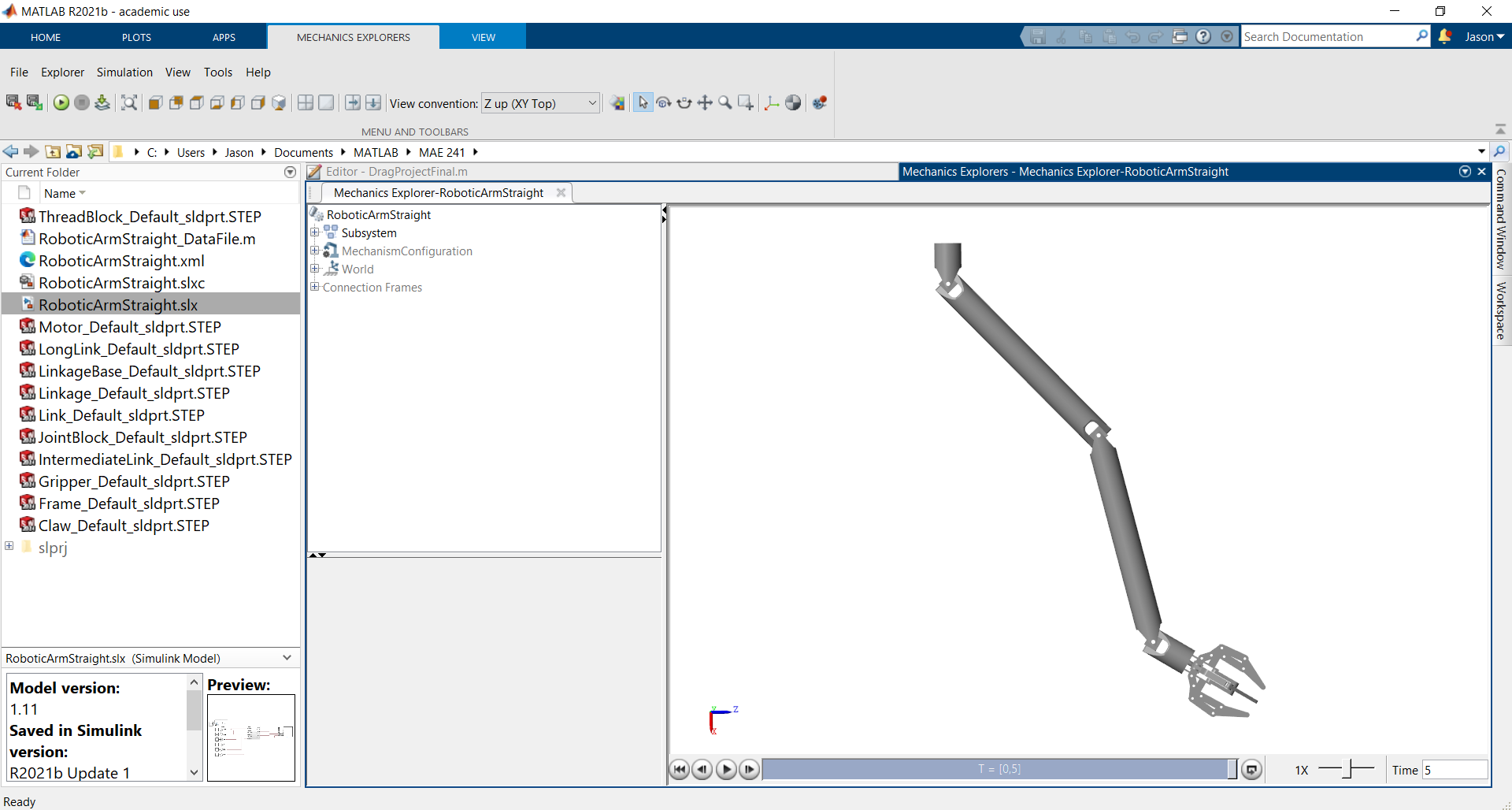}
    \caption{The 3D simulation environment (Mechanics Explorer) in MATLAB} 
    \vspace{-5mm}
    \label{fig:MechanicsExplorer}
\end{figure}

A simulation environment called the ``Mechanics Explorer,'' shown in Fig. \ref{fig:MechanicsExplorer}, displays the 3D model from many different viewpoints. 

\subsection{Integration with Drone}

An existing drone model ``Quadcopter Payload Delivery'' from MATLAB examples was used for the integration of the SolidWorks model of the robot arm. This approach allows efficient combination of various components to achieve a fully functional and sophisticated system.

\begin{figure}[t!]
    \centering
    \includegraphics[width=0.3\textwidth]{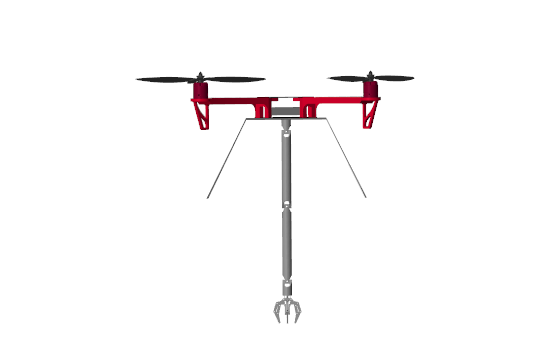}
    \caption{Drone with a robot arm}
    \label{fig:Quadcopter-arm}
\end{figure}

\begin{figure}[t!]
    \centering
    \includegraphics[width=0.5\textwidth]{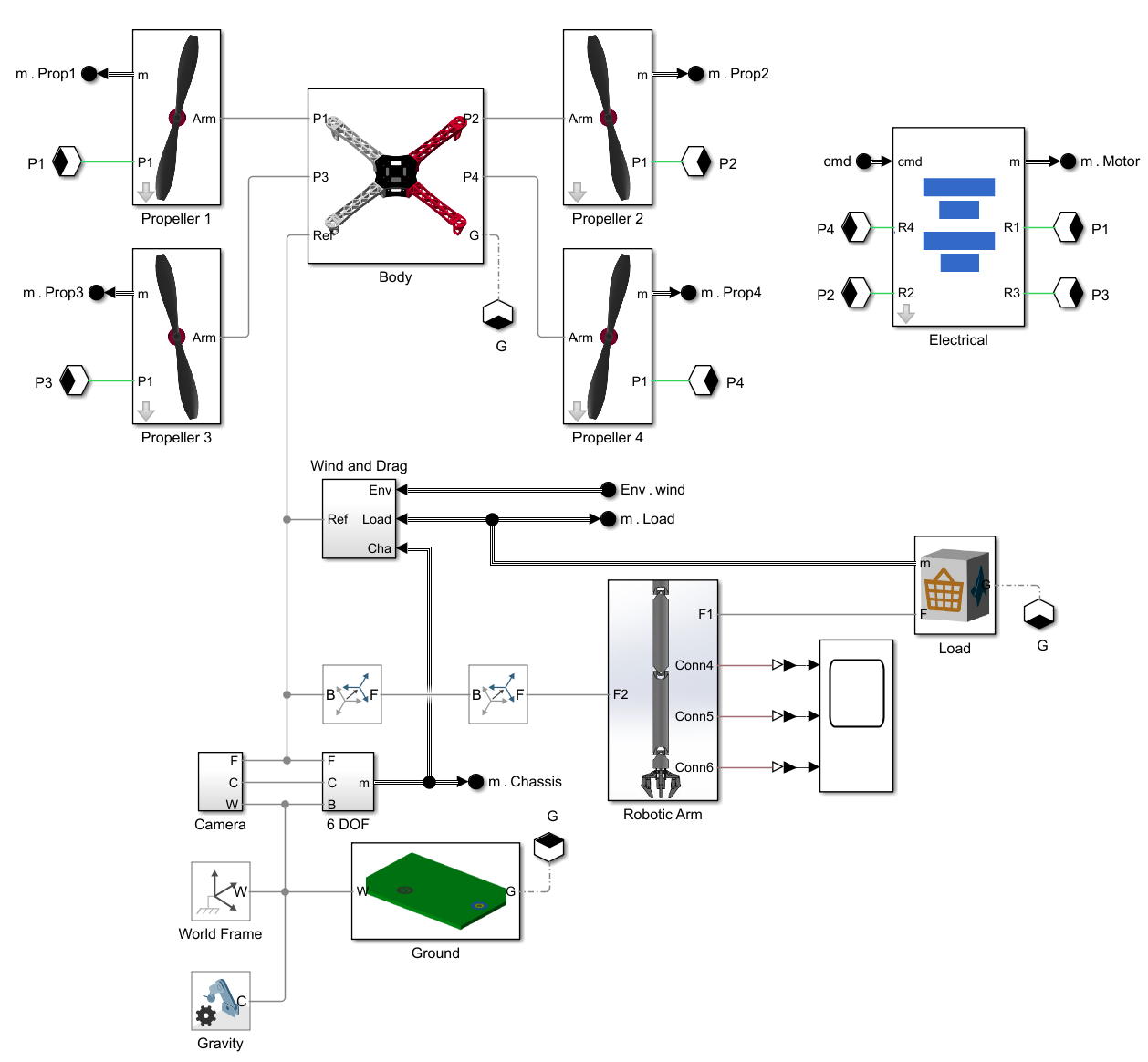}
    \caption{Modified Quadcopter Block with the arm attached as a subsystem}
    \vspace{-5mm}
    \label{fig:Quadarm-block}
\end{figure}

The generated Simscape model contains four main blocks on the surface: the Reference trajectory block, the Maneuver Controller, the Quadcopter, robotic arm block, and payload block, and the Scope block (see Fig. \ref{fig:Quadarm-block} and \ref{fig:Quadcopter-arm}).

\subsection{Flight Control}
\begin{figure}[t!]
    \centering
    \includegraphics[width=0.5\textwidth]{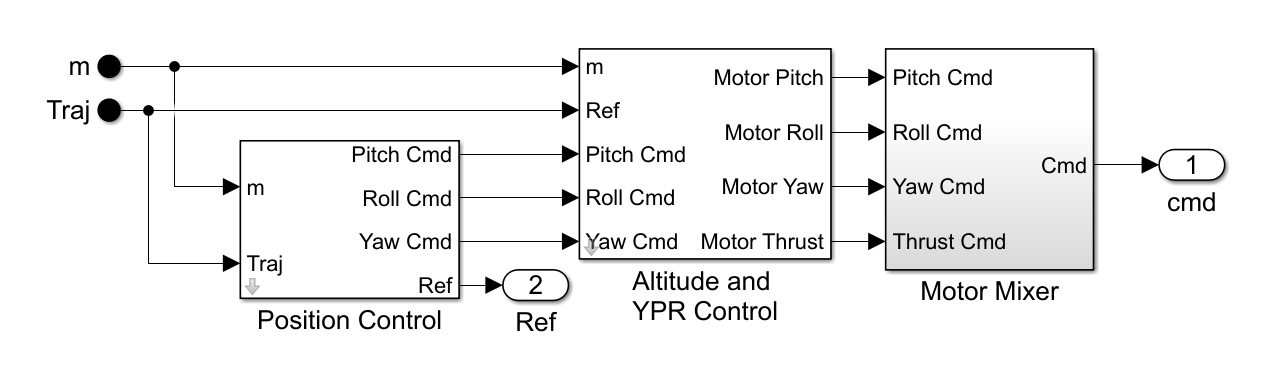}
    \caption{Maneuver controller block expanded view}
    \vspace{-5mm}
    \label{fig:manuever_controller}
\end{figure}

The Maneuver Controller Block in Fig. \ref{fig:manuever_controller} is responsible for controlling the entire drone behaviour, its yaw, pitch, roll and thrust corresponding to reference values. In general drone controllers use a two loop control design, wherein the outer loop minimises the position error of the drone and the inner loop is responsible for the flight attitude and thrust generated by the motors. In this design the Position Control block does all the outer loop calculations and the Altitude and YPR Control Block does the inner loop calculations which is then fed into the Motor Mixing Block which decides which motor and propeller should produce how much thrust to achieve the desired orientation and position.
Two approaches were planned for the design of the maneuver controller block, first the existing model from the example used PIDs for position and orientation and second using a Model Reference Adaptive Controller for Yaw, Pitch and Roll controller, with a conventional PID tuned for thrust control.

\subsubsection{PID Based Controller}
A PID controller $u$ design involves adjusting the proportional ($K_P$), integral ($K_I$), derivative ($K_D$) gains, and Filter Coefficient ($N$) in (\ref{eq:PID}) to get the desired system response. 
\begin{equation}
    U(s) = K_p + K_I \frac{1}{s} + K_D \frac{N}{1 + N \frac{1}{s}}
    \label{eq:PID}
\end{equation}
The following controller parameters as in Table \ref{tab:PID Gains} have been used for the thrust, roll, pitch and yaw controllers:
\begin{table}[H]
\begin{center}
\caption{Designed PID Controller Gains}
    \begin{tabular}{|c|c|c|c|c|}
    \hline
       Controller & $K_P$ & $K_I$ & $K_D$ & N\\
       \hline
        Thrust & 0.25 & 0.05 & 0.35 & 10000\\
        \hline
        Roll & 100 & 0 & 800 & 1000\\
        \hline
        Pitch & 100 & 0 & 800 & 1000\\
        \hline
        Yaw & 205.61 & 0.059203 & 0.782 & 100\\
        \hline
    \end{tabular}
    \label{tab:PID Gains}
\end{center}
\end{table}
\vspace{-15pt}
We used Ziegler-Nichols Method to determine the initial gains followed by manual tuning to get acceptable performance and minimize the trajectory tracing error.

\subsubsection{ Model Reference Adaptive Control}
Model Reference Adaptive Control (MRAC) is an adaptive control strategy designed to automatically adjust the parameters of a controller so that the behavior of the controlled system matches that of a reference model, even in the presence of system uncertainties or changing dynamics \cite{MRAC2018}. The primary goal of MRAC is to ensure that the output of the uncertain system follows the desired reference model's output.
\begin{comment}
\begin{figure}[!h]
    \centering
    \includegraphics[width=0.48\textwidth]{Figures/Direct_MRAC.png}
    \caption{Direct MRAC block diagram.}
    \label{fig:DirectMRAC}
\end{figure}
\end{comment}
Direct MRAC is a specific type of MRAC where control gains are adjusted in real-time to minimize the tracking error—the difference between the system's output and the reference model's output—without needing to estimate the system’s parameters directly \cite{MRACNguyen2018}. This is especially useful in scenarios with significant uncertainties in system dynamics or in situations with unknown or time-varying system parameters. The control signal in direct MRAC is typically expressed as a function of the system states and reference commands, with time-varying control gains. The control law can be written as in (\ref{eq:MRAC})
\begin{equation}
u(t) = k_x(t) x(t) + k_r(t) r(t) - \Theta(t)^T \Phi(x)
\label{eq:MRAC}
\end{equation}
where $x(t)$ is the system state, $r(t)$ is the reference input, and $\Phi(x)$ is a vector of known basis functions representing structured uncertainties. The adaptation process aims to drive the control gains towards their ideal values, which would ensure zero tracking error.

The adaptation mechanism in direct MRAC is derived from stability considerations using Lyapunov's theory. A candidate Lyapunov function incorporating the tracking error and estimation errors is chosen, and its time derivative is evaluated to guarantee system stability. For instance, in a first-order system, the Lyapunov function can be defined as:
\[
V(e, k_x, k_r, \Theta) = e^2 + \frac{|b|}{\gamma_x} k_x^2 + \frac{|b|}{\gamma_r} k_r^2 + \Theta^T \Gamma^{-1} \Theta,
\]
where $e(t)$ is the tracking error, and $\gamma_x$, $\gamma_r$, and $\Gamma$ are adaptation rate parameters. The control gains are updated using the following adaptive laws:
\[
\dot{k}_x = \gamma_x x e^T P B,  \quad  \dot{k}_r = \gamma_r r e^T P B, \quad  \dot{\Theta} = \Gamma \Phi(x) e^T P B,
\]
where $P$ is the solution to the following Lyapunov equation, $A_m^T P + P A_m + Q = 0$, and $Q$ is a positive definite square matrix of size state vector $x(t)$. 
These update laws ensure that the tracking error approaches zero as time progresses, leading to asymptotic tracking of the reference model. While the tracking error converges to zero, the adaptive parameters (such as the control gains) are generally only bounded, not necessarily converging to constant values.

The MRAC method consists of three main elements: feedback, feedforward, and adaptive control. The feedback and feedforward gains are adjusted online using a learning rate denoted as $\gamma_k = 110$. A single hidden layer (SHL) neural network is used to estimate and cancels model uncertainties online. The SHL network features the following parameters:
\begin{itemize}
\item Outer Layer Learning Rate ($\gamma_w$) = 5
\item Inner-layer Learning Rate ($\gamma_v$) = 1
\item Number of SHL Neurons (N) = 50
\item Lyapunov coefficient for controller weight updates $Q = [125, 200, 125, 200, 120, 125]$
\end{itemize}

\section{Results} \label{chp:results}
\begin{figure}[t!]
    \centering
    \includegraphics[width=0.46\textwidth, height = 0.4\textwidth]{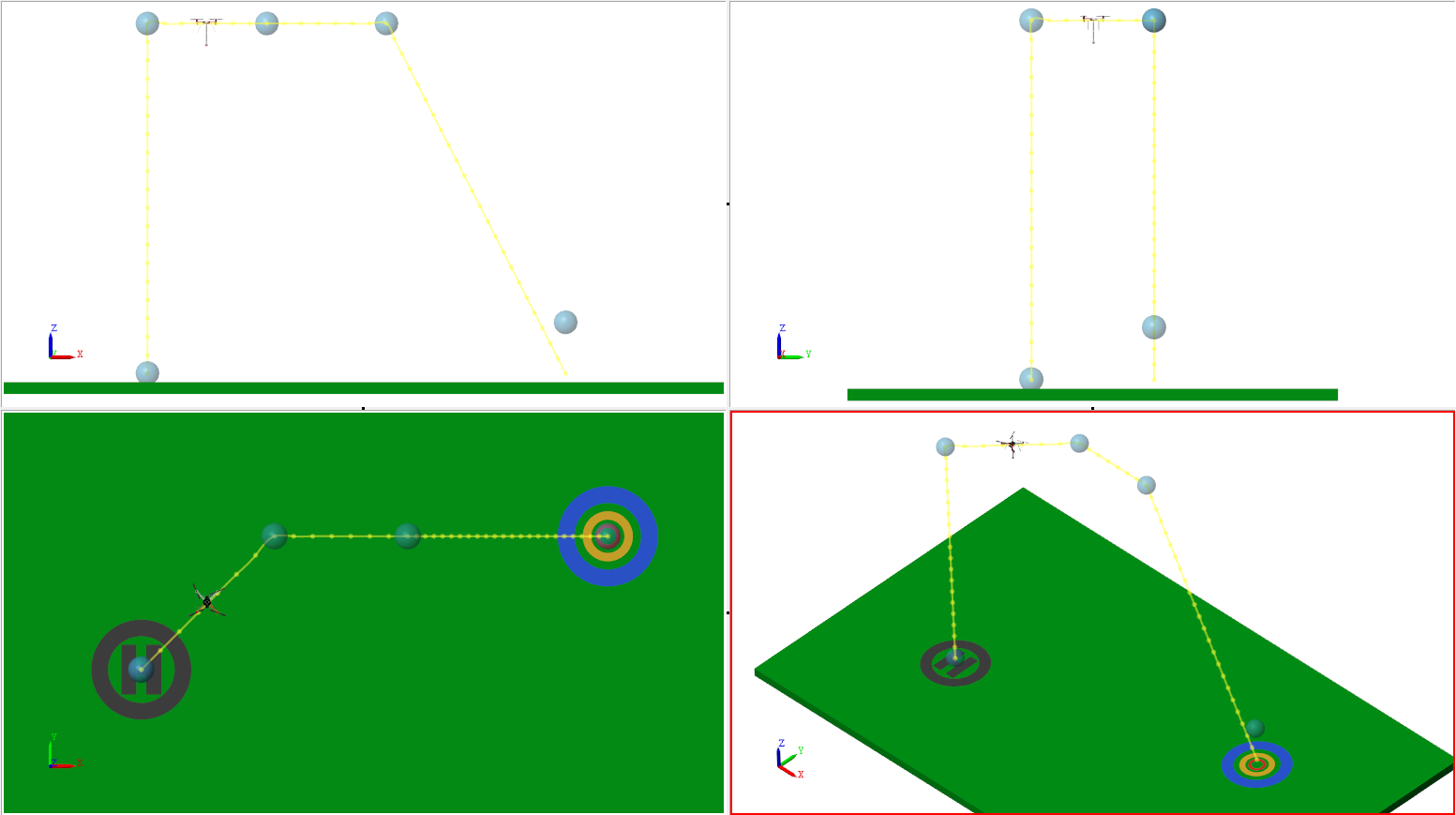}
    \caption{Reference trajectory tracking}
    \vspace{-3mm}
    \label{fig:Four_view}
\end{figure}

The performance of each controller is evaluated on their ability to track the desired trajectory as close as possible. The simulation results for tracking a desired trajectory, designed using Waypoints have been shown in Fig. \ref{fig:Four_view}.

\begin{figure}[t!]
    \centering
    \includegraphics[width=0.48\textwidth, height = 0.45\textwidth]{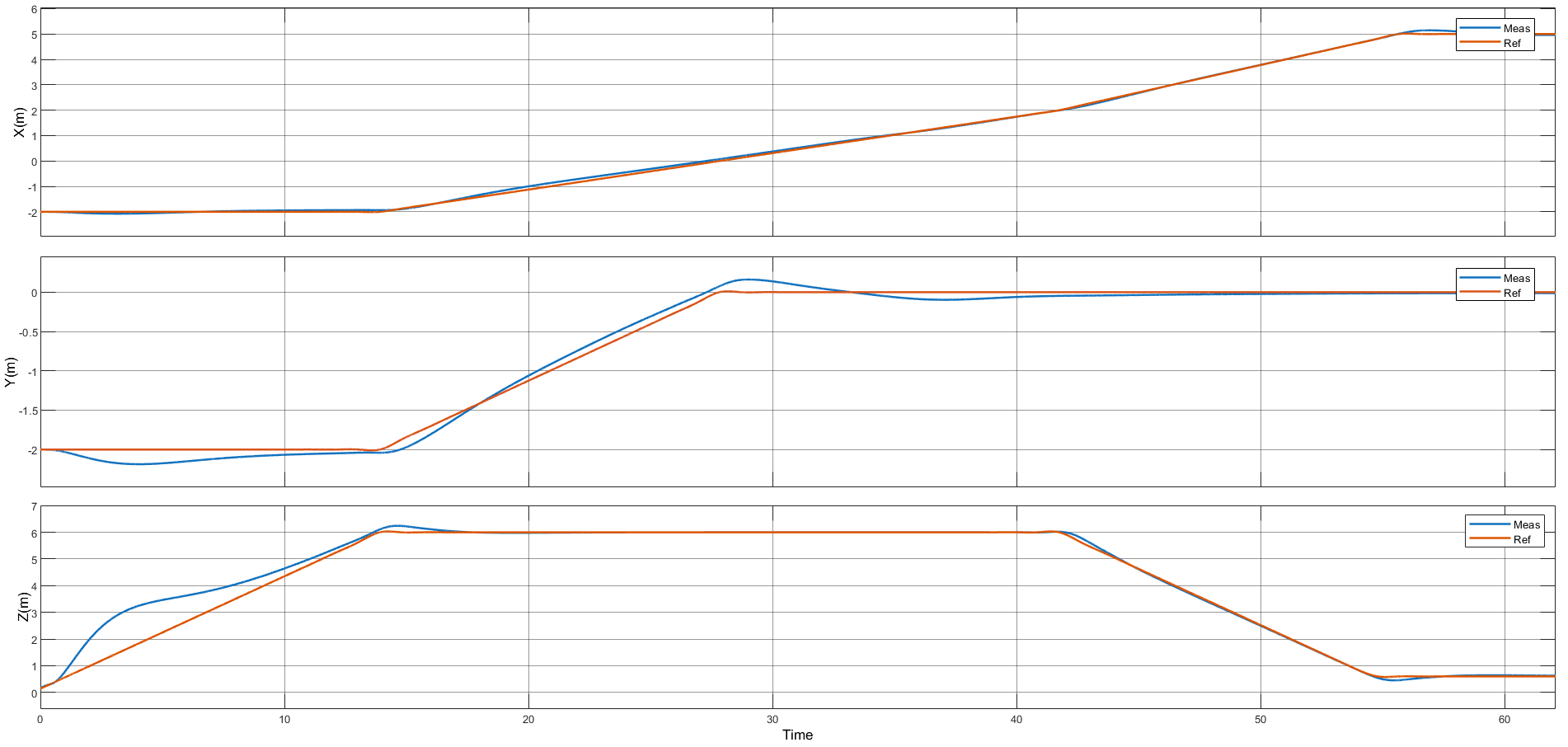}
    \caption{Trajectory tracking with Package with PID Controllers}
    \vspace{-3mm}
    \label{fig:PID_result_Pack}
\end{figure}

\begin{figure}[t!]
    \centering
    \includegraphics[width=0.48\textwidth, height = 0.48\textwidth]{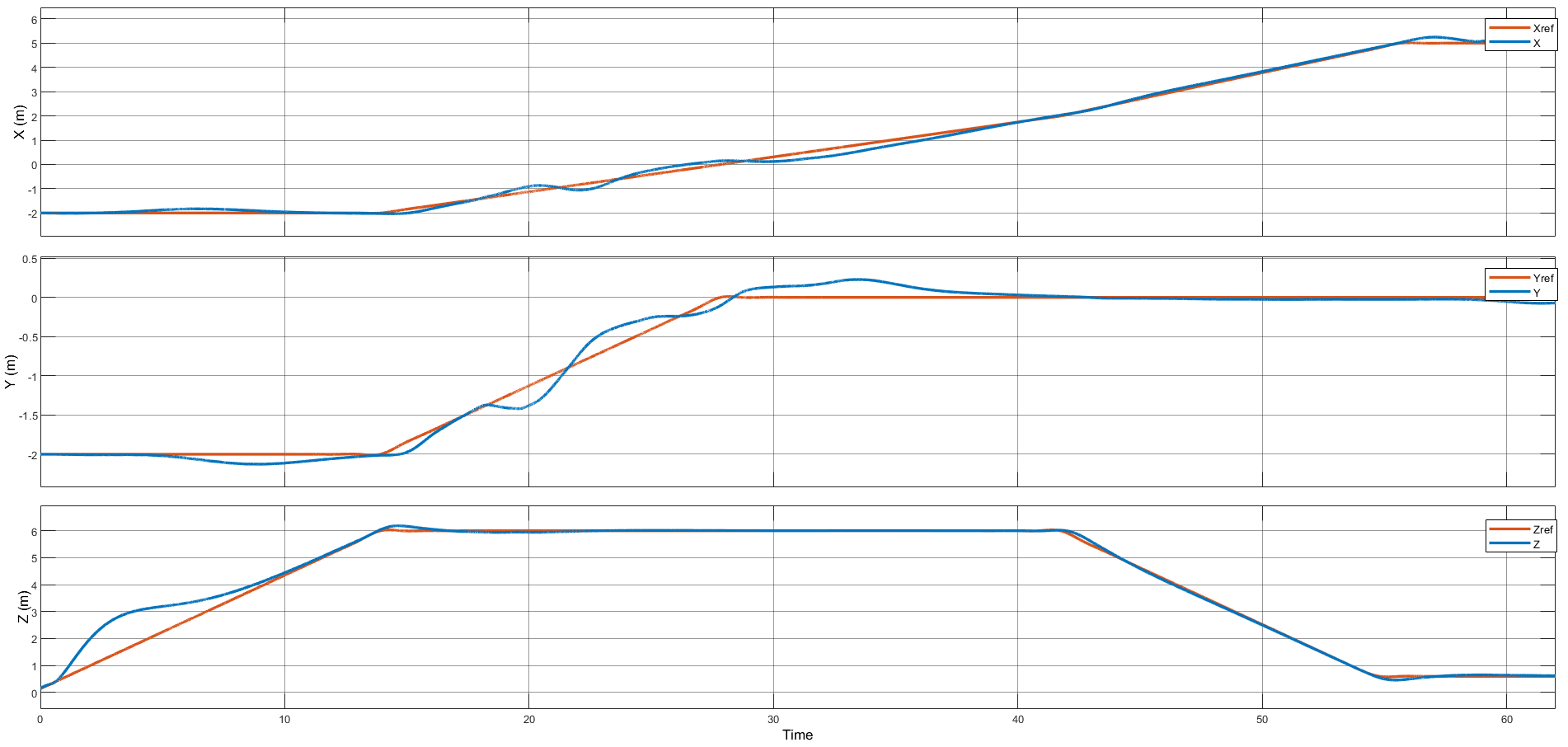}
    \caption{Trajectory tracking with Package with MRAC Controller}
    \vspace{-5mm}
    \label{fig:MRAC_result_Pack}
\end{figure}

As evident from Fig. \ref{fig:PID_result_Pack} and \ref{fig:MRAC_result_Pack}, PID controllers track the reference trajectory more accurately. There is less overshoot from the reference values in case of the PID controller. Thus, it may also be worthwhile to use the MRAC in conjunction with PID, as opposed to having a seperate PID Controller for the thrust, as shown by Dan Zhang and Bin Wei \cite{ZhangUnknownTitle2016}, which may make it possible to have a more tuned MRAC controller. The RMS errors for cases when a payload is attached and not attached are given in Table \ref{tab:RMS Errors}. The PID controller is a better controller as it has lower RMS errors throughout the entire trajectory.
\begin{table}[H]
\begin{center}
\caption{RMS Errors}
\scriptsize
\begin{tabular}{|p{1.2cm}|p{0.8cm}p{0.3cm}|p{0.8cm}p{0.3cm}|p{0.8cm}p{0.3cm}|}
\hline
\multicolumn{1}{|c|}{\textbf{Error}} & \multicolumn{2}{c|}{\textbf{X Error}} & \multicolumn{2}{c|}{\textbf{Y Error}} & \multicolumn{2}{c|}{\textbf{Z Error}} \\ \hline
Controller  & \multicolumn{1}{c|}{PID} & \multicolumn{1}{c|}{MRAC} & \multicolumn{1}{c|}{PID} & \multicolumn{1}{c|}{MRAC} & \multicolumn{1}{c|}{PID} & \multicolumn{1}{c|}{MRAC} \\ \hline
Payload     & \multicolumn{1}{l|}{0.037} & 0.496  & \multicolumn{1}{l|}{0.037} & 0.071  & \multicolumn{1}{l|}{0.012} & 0.021 \\ \hline
No Payload  & \multicolumn{1}{l|}{0.038} & 0.485  & \multicolumn{1}{l|}{0.038} & 0.327  & \multicolumn{1}{l|}{0.011} & 0.021 \\ \hline
\end{tabular}
    \label{tab:RMS Errors}
\end{center}
\end{table}

\section{Conclusion} \label{chp:conclusion}
This study undertakes a thorough exploration the physical modeling and integration of a robotic arm with a quadcopter, utilizing SolidWorks for mechanical design and MATLAB Simscape for simulation to create a reliable and efficient simulator that mimics real-world dynamics. The paper provides the details of the design of the robotic arm, its integration with the drone, and design of two control strategies for trajectory tracking. 

The development process highlighted several key challenges: ensuring mechanical compatibility between the robotic arm and the quadcopter, maintaining stability during flight, and accurately simulating the system's physical interactions.

Control strategies using PID and MRAC were designed and rigorously tested. While PID controllers demonstrated superior trajectory tracking and stability, they required extensive tuning efforts. MRAC offered adaptability to changing dynamics but exhibited higher RMS errors.

The results underscore the importance of reliable physical modeling in achieving accurate simulation outcomes. The combined use of SolidWorks and MATLAB Simscape facilitated the creation of a simulator that not only addresses the physical and mechanical challenges but also aids in the design and tuning of high-reliability controllers. The process from formulating the equations of motion to developing a control scheme represents a comprehensive exploration into the advanced dynamics of robotic arms attached to drones.

Future work will focus on further refining control algorithms and exploring real-world applications to validate the simulator's effectiveness in practical scenarios. This project not only contributes to the theoretical understanding of such systems but also lays the groundwork for practical advancements with tangible applications in robotics and aerial systems.

\addtolength{\textheight}{-10cm}   % This command serves to balance the column lengths

%%%%%%%%%%%%%%%%%%%%%%%%%%%%%%%%%%%%%%%%%%%%%%%%%%%%%%%%%%%%%%%%%%%%%%%%%%%%%%%%

%%%%%%%%%%%%%%%%%%%%%%%%%%%%%%%%%%%%%%%%%%%%%%%%%%%%%%%%%%%%%%%%%%%%%%%%%%%%%%%%

%%%%%%%%%%%%%%%%%%%%%%%%%%%%%%%%%%%%%%%%%%%%%%%%%%%%%%%%%%%%%%%%%%%%%%%%%%%%%%%%

\bibliographystyle{IEEEtran}
\bibliography{references}

\begin{thebibliography}{10}
\providecommand{\url}[1]{#1}
\csname url@rmstyle\endcsname
\providecommand{\newblock}{\relax}
\providecommand{\bibinfo}[2]{#2}
\providecommand\BIBentrySTDinterwordspacing{\spaceskip=0pt\relax}
\providecommand\BIBentryALTinterwordstretchfactor{4}
\providecommand\BIBentryALTinterwordspacing{\spaceskip=\fontdimen2\font plus
\BIBentryALTinterwordstretchfactor\fontdimen3\font minus \fontdimen4\font\relax}
\providecommand\BIBforeignlanguage[2]{{%
\expandafter\ifx\csname l@#1\endcsname\relax
\typeout{** WARNING: IEEEtran.bst: No hyphenation pattern has been}%
\typeout{** loaded for the language `#1'. Using the pattern for}%
\typeout{** the default language instead.}%
\else
\language=\csname l@#1\endcsname
\fi
#2}}

\bibitem{article}
D.~Cekus, B.~Posiadala, and P.~Waryś, ``Integration of modeling in solidworks and matlab/simulink environments,'' \emph{Archive of Mechanical Engineering}, vol.~61, 03 2014.

\bibitem{article2}
M.~Pozzi, G.~Achilli, M.~Valigi, and M.~Malvezzi, ``Modeling and simulation of robotic grasping in simulink through simscape multibody,'' \emph{Frontiers in Robotics and AI}, vol.~9, p. 873558, 05 2022.

\bibitem{Jatsun2020SynthesisOS}
S.~F. Jatsun, B.~Lushnikov, O.~Emelyanova, and A.~S.~M. Leon, ``Synthesis of simmechanics model of quadcopter using solidworks cad translator function,'' in \emph{Proceedings of 15th International Conference on Electromechanics and Robotics ``Zavalishin's Readings''}, 2020.

\bibitem{mahto2022performance}
R.~K. Mahto, J.~Kaur, and P.~Jain, ``Performance analysis of robotic arm using simulink,'' in \emph{2022 IEEE World Conference on Applied Intelligence and Computing (AIC)}, 2022, pp. 508--512.

\bibitem{Long2020}
D.~T. Long, T.~V. Binh, R.~V. Hoa, L.~V. Anh, and N.~V. Toan, ``Robotic arm simulation by using matlab and robotics toolbox for industry application,'' \emph{SSRG International Journal of Electronics and Communication Engineering}, vol.~7, no.~10, pp. 1--4, 2020.

\bibitem{Garcia2021}
M.~Garcia, P.~Pena, A.~Tekes, and A.~A. Amiri~Moghadam, ``Development of novel three-dimensional soft parallel robot,'' in \emph{SoutheastCon 2021}, 2021, pp. 1--6.

\bibitem{pena2020}
P.~Pena, M.~Garcia, and A.~Tekes, ``Modeling of compliant mechanisms in matlab simscape,'' in \emph{ASME International Mechanical Engineering Congress and Exposition}, vol. 84553.\hskip 1em plus 0.5em minus 0.4em\relax American Society of Mechanical Engineers, 2020, p. V07BT07A020.

\bibitem{Lee2018}
K.~Lee, J.~Lee, B.~Woo, and J.~Lee, ``Modeling and control of an articulated robot arm with embedded joint actuators,'' in \emph{2018 International Conference on Information and Communication Technology Robotics (ICT-ROBOT)}, 2018, pp. 1--4.

\bibitem{icscan2018design}
M.~{\.I}{\c{s}}can, H.~Eken, B.~Vural, and C.~Y{\i}lmaz, ``Design and control of an exoskeleton robot: A matlab simscape application,'' \emph{J. Therm. Eng}, vol.~4, pp. 1867--1878, 2018.

\bibitem{Urrea16}
C.~Urrea, L.~Valenzuela, and J.~Kern, ``Design, simulation, and control of a hexapod robot in simscape multibody,'' in \emph{Applications from Engineering with MATLAB Concepts}, J.~Valdman, Ed.\hskip 1em plus 0.5em minus 0.4em\relax Rijeka: IntechOpen, 2016, ch.~5.

\bibitem{Urrea2020}
C.~Urrea and D.~Saa, ``Design and implementation of a graphic simulator for calculating the inverse kinematics of a redundant planar manipulator robot,'' \emph{Applied Sciences}, vol.~10, no.~19, p. 6770, 2020.

\bibitem{Eldirdiry2020}
O.~Eldirdiry, R.~Zaier, A.~Al-Yahmedi, I.~Bahadur, and F.~Alnajjar, ``Modeling of a biped robot for investigating foot drop using matlab/simulink,'' \emph{Simulation Modelling Practice and Theory}, vol.~98, p. 101972, 2020.

\bibitem{Guedelha2022}
N.~Guedelha, V.~Pasandi, G.~L’Erario, S.~Traversaro, and D.~Pucci, ``A flexible matlab/simulink simulator for robotic floating-base systems in contact with the ground: Theoretical background and implementation details,'' \emph{International Journal of Semantic Computing}, vol.~18, no.~02, pp. 239--255, 2024.

\bibitem{Yura2017}
J.~Yura, M.~Oyun-Erdene, B.~E. Byambasuren, and D.~Kim, ``Modeling of violin playing robot arm with matlab/simulink,'' in \emph{Robot Intelligence Technology and Applications 4}, ser. Advances in Intelligent Systems and Computing.\hskip 1em plus 0.5em minus 0.4em\relax Springer, Cham, 2017, vol. 447, pp. 167--176.

\bibitem{Zhang2019}
C.~Zhang and Z.~Zhang, ``Modelling and simulation of scara robot using matlab/simmechanics,'' in \emph{2019 IEEE 3rd Advanced Information Management, Communicates, Electronic and Automation Control Conference (IMCEC)}, 2019, pp. 516--519.

\bibitem{niku2020introduction}
S.~B. Niku, \emph{Introduction to robotics: analysis, control, applications}.\hskip 1em plus 0.5em minus 0.4em\relax John Wiley \& Sons, 2020.

\bibitem{MRAC2018}
A.~Shekhar and A.~Sharma, ``Review of model reference adaptive control,'' in \emph{2018 International Conference on Information , Communication, Engineering and Technology (ICICET)}, 2018, pp. 1--5.

\bibitem{MRACNguyen2018}
N.~T. Nguyen, \emph{Model-Reference Adaptive Control}.\hskip 1em plus 0.5em minus 0.4em\relax Cham: Springer International Publishing, 2018, pp. 83--123.

\bibitem{ZhangUnknownTitle2016}
D.~Zhang and B.~Wei, ``Convergence performance comparisons of pid, mrac, and pid + mrac hybrid controller,'' \emph{Frontiers of Mechanical Engineering}, vol.~11, pp. 213--217, MAY 2016.

\end{thebibliography}

\end{document}